\documentclass{article}
\usepackage{ijcai23}

\usepackage{times}
\usepackage{soul}
\usepackage{url}
\usepackage[hidelinks]{hyperref}
\usepackage[utf8]{inputenc}
\usepackage[small]{caption}
\usepackage{graphicx}
\usepackage{amsmath}
\usepackage{amsthm}
\usepackage{booktabs}
\usepackage{algorithm}
\usepackage{algpseudocode}
\usepackage[switch]{lineno}
\usepackage{natbib}

\usepackage[dvipsnames]{xcolor}
\usepackage{comment}
\usepackage{float}
\usepackage{amsthm}
\usepackage{amssymb}
\usepackage{mathtools}
\usepackage{algorithm}
\usepackage{float}
\usepackage{centernot} 
\usepackage{newunicodechar}
\usepackage{changepage}
\DeclareMathSymbol{,}{\mathalpha}{operators}{"2C} 

\newunicodechar{∪}{\ensuremath{\cup}}

\DeclareMathOperator{\vars}{\mathit{vars}}
\DeclareMathOperator{\pre}{\mathit{pre}}
\DeclareMathOperator{\eff}{\mathit{eff}}
\DeclareMathOperator{\cost}{\mathit{c}}

\newtheorem{theorem}{Theorem}
\newtheorem{lemma}{Lemma}
\newtheorem{definition}{Definition}

\title{Task Scoping: Generating Task-Specific Abstractions\\ for Planning in Open-Scope Models}

\author{
Michael Fishman\thanks{Equal contribution}\and
Nishanth Kumar\footnotemark[1]$^1$\and
Cameron Allen$^2$\and 
Natasha Danas$^2$\and
Michael Littman$^2$\and
Stefanie Tellex$^2$\And
George Konidaris$^2$\\
\affiliations
$^1$MIT CSAIL, $^2$Brown University Department of Computer Science\\
\emails
michael@fishman.ai, njk@csail.mit.edu
}

\begin{document}

\maketitle

\begin{abstract}
A general-purpose planning agent requires an open-scope world model: one rich enough to tackle any of the wide range of tasks it may be asked to solve over its operational lifetime. This stands in contrast with typical planning approaches, where the scope of a model is limited to a specific family of tasks that share significant structure. Unfortunately, planning to solve any specific task using an open-scope model is computationally intractable---even for state-of-the-art methods---due to the many states and actions that are necessarily present in the model but irrelevant to that problem. We propose task scoping: a method that exploits knowledge of the initial state, goal conditions, and transition system to automatically and efficiently remove provably irrelevant variables and actions from a planning problem. Our approach leverages causal link analysis and backwards reachability over state variables (rather than states) along with operator merging (when effects on relevant variables are identical). Using task scoping as a pre-planning step can shrink the search space by orders of magnitude and dramatically decrease planning time. We empirically demonstrate that these improvements occur across a variety of open-scope domains, including Minecraft, where our approach leads to a $75\times$ reduction in search time with a state-of-the-art numeric planner, even after including the time required for task scoping itself.
\end{abstract}
\section{Introduction}
Modern AI planning is extremely general-purpose---the promise of domain-independent planners is that a single program can be used to solve planning tasks arising from many specific applications.
This promise has largely been realized:
given an appropriately specified model of a planning problem, modern planners can quickly tackle anything from game playing \citep{KORF198597, KORF198535} to transportation logistics \citep{planning_for_transportation} to chemical synthesis \citep{Matloob2016ExploringOS}.

But while the achievements of domain-independent planners are impressive, one pervasive assumption significantly limits their generality: namely, that the domain model is always well-matched to the task.
The problem encodings for all the applications discussed above were carefully designed by human experts.
While each domain supports multiple problems, these problems all share significant structure; a planner cannot solve logistics problems with a chemical synthesis model.
The model has limited \emph{scope}; it is only intended to be compatible with a specific, restricted set of possible tasks.

\begin{figure}
    \centering\includegraphics[width=8.0cm,clip]{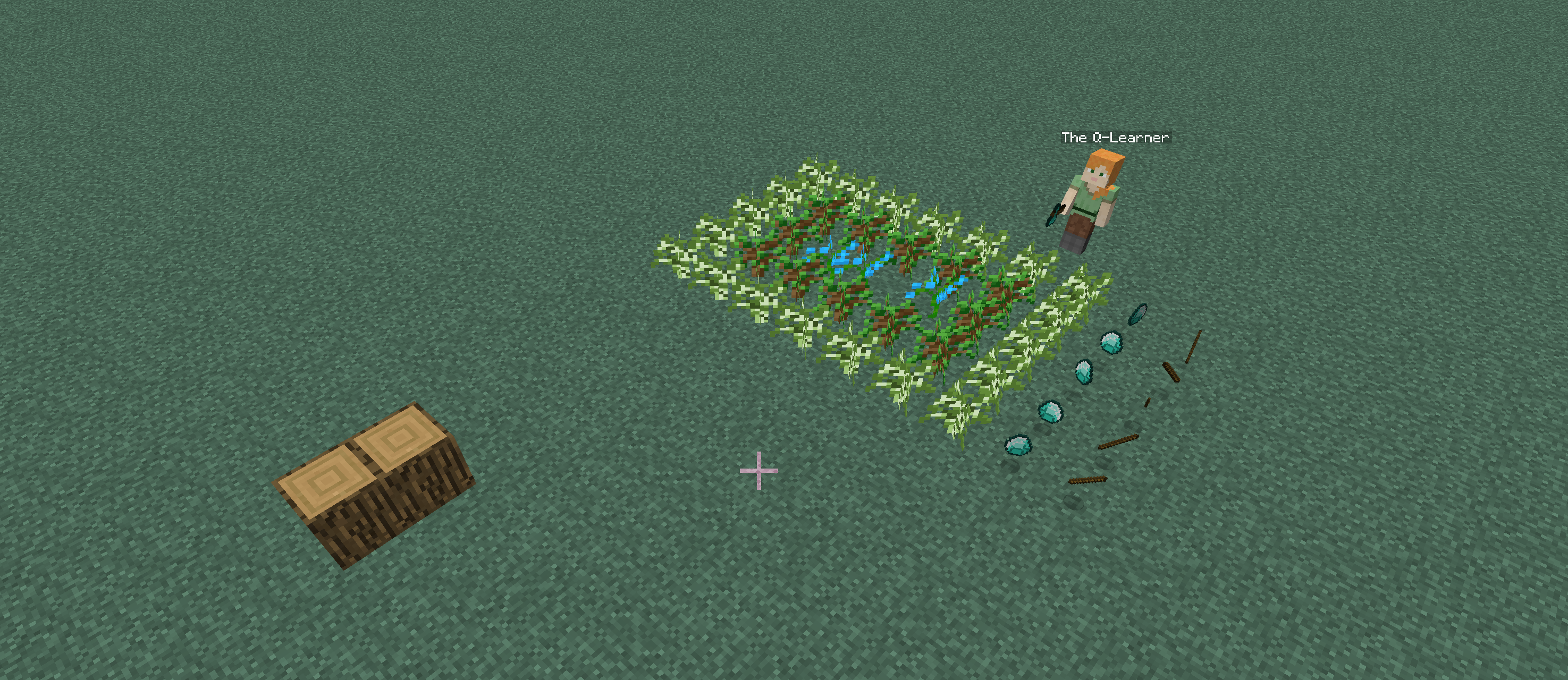}
    \caption{\small{An open-scope Minecraft environment. All of these objects are occasionally important; however, for the specific task of crafting a bed, the planning agent can ignore most of them. Task scoping removes irrelevant objects and actions from the domain prior to planning, reducing planning time by an order of magnitude.}}
    \label{fig:FrontPage}
\end{figure}

By contrast, general-purpose planning agents cannot assume there will be a human expert on hand to provide them with a carefully specified model for any problem they might face.
Instead, they must maintain an \emph{open-scope} model---one rich enough to describe any planning task they may encounter during their deployment.
Such an open-scope model will necessarily contain large amounts of information irrelevant to any individual task \citep{GeorgeNecessity}.
For instance, a Minecraft planning agent must possess a model capable of expressing tasks such as obtaining resources, building shelter, crafting weapons, cooking food, and fighting off enemies---to name a few.
However, when confronted with the specific, immediate task of crafting a bed, information about other tasks like cooking food is simply irrelevant.

Unfortunately, that generality comes at a cost: when models contain large amounts of irrelevant information, the search space grows exponentially and planning quickly becomes intractable.
Recent work \citep{VallatiPlanningRobustness,TomAndRohanPLOI} has shown that many state-of-the-art planning engines
suffer significant reductions in performance when irrelevant objects, state variables, or operators  are included in domain descriptions.
For example, the state-of-the-art numeric planner ENHSP-2020 \citep{Scala2020SubgoalingTF} fails to find an optimal plan to construct a bed from the objects in the Minecraft domain of Figure \ref{fig:FrontPage}.
Fast Downward \citep{Helmert:2006:FDP:1622559.1622565} fails to even translate a non-numeric version of the same problem.
With irrelevant objects and actions removed, both planners can solve the task within a few minutes.

To help general-purpose planning agents overcome these challenges, we introduce \emph{task scoping}, a method for reasoning about which information can safely be removed from the agent's model.
We identify three types of task-scoping abstractions that agents can use, prior to planning, to remove irrelevant actions and variables.
First, agents need only consider actions to be relevant if they modify goal variables or preconditions of other relevant actions.
Second, agents can identify actions that have identical effects over relevant variables, and merge them.
Third, agents can additionally ignore variables that already match relevant preconditions and goal clauses, unless a relevant action can modify them.

We prove that task scoping preserves all optimal plans and empirically demonstrate that it leads to substantial reductions in search space size and planning time.
Applying task scoping to the Minecraft problem of Figure \ref{fig:FrontPage}
allows us to solve the previously intractable problem in under $3$ minutes.
We also observe significant improvements on a variety of other numeric and classical planning domains.
Most importantly, the entire process is automatic, and compatible with off-the-shelf planners, enabling agents to build their own task-specific abstractions for planning with open-scope models, all without requiring any additional domain knowledge from human experts.

\section{Background}
We define a planning domain (or model) in terms of the following quantities.
\begin{itemize}
    \item A finite set of variables $\mathcal{V}$, where each variable $v$ can take on values from some domain $\mathcal{D}(v)$. Our approach is compatible with variables that are binary propositions, enums, or numeric fluents, so $\mathcal{D}(v)$ can be either finite or infinite.
    \item A factored state space $\mathcal{S}$ comprised of the state variables in $\mathcal{V}$. A state $s$ assigns a value to every variable in $\mathcal{V}$.
    \item A set of grounded operators (or actions) $\mathcal{O}$, where each operator $o$ consists of:
    \begin{itemize}
        \item A precondition $\pre(o)$, consisting of a logical expression defined over variables in $\mathcal{V}$;
        \item An effect $\eff(o)$, consisting of an assignment of values (potentially determined by a function, such as increment/decrement) to some or all of the variables in $\mathcal{V}$;
        \item A cost $\cost(o) > 0$.
    \end{itemize}
\end{itemize}
An operator $o$ is applicable in state $s$ if $s$ implies $\pre(o)$. Executing $o$ incurs some cost $c(o)$ and causes the variable assignments described in $\eff(o)$, resulting in a new state $s'$. We will also frequently discuss the variables appearing in $\pre(o)$ and $\eff(o)$ using the notation $\vars(\pre(o))$ and $\vars(\eff(o))$.\footnote{If a function assigns a value to some variable $v$, such that the assignment depends on another variable $u$, then $u$ is considered a precondition variable of the operator, despite not appearing in the precondition.}

We define a problem instance (or task) of a given domain by adding the following additional quantities:
\begin{itemize}
    \item An initial state $s_0 \in \mathcal{S}$;
    \item A goal condition $G$ consisting of an assignment to some (or all) of the variables in $\mathcal{V}$.
\end{itemize}
A valid plan is a sequence of successively applicable operators $[o_1, o_2, ..., o_n]$ from initial state $s_0$ to final state $s_n$, such that $s_n$ implies $G$. An optimal plan is any valid plan that incurs the minimum cost. Given a model, a task is \emph{solvable} if there exists at least one valid plan using that model.

This formalism is compatible with a variety of planning problem encodings, including PDDL 2.1 level 2 \citep{PDDL2.1} and SAS+  \citep{Sandewall1986ARO}. In our experiments, we support both the former without conditional effects and the latter without axioms, although each of these restrictions could easily be relaxed in future work.

\subsection{Open-Scope Models}
\label{sec:open-scope-models}
A general-purpose planning agent may be asked solve many tasks during its operational lifetime. If the agent's model contains too few variables or operators, some tasks will not be solvable. To ensure that as many potential tasks as possible are solvable, the agent's model must be open-scope: it must contain more variables and/or operators than are relevant for any one task.

Given a model $M$ and a task $t$, we say that an operator is \emph{task-relevant} (or simply \emph{relevant}) if it appears in at least one optimal plan to solve $t$.
We say that $M$ is an \emph{open-scope} model with respect to $t$ if it contains operators that are not relevant (henceforth `irrelevant'). The wider the range of tasks an agent may be asked to solve, the higher the likelihood that most of its operators will be irrelevant for any one of those tasks. These irrelevant operators are the ones we would like the agent to ignore with task scoping.

\subsubsection{Example}
Consider the following simplified version of the Minecraft domain in Figure \ref{fig:FrontPage}. The agent can collect sticks, food, and stone, which it needs for eating and making an axe. The task is to make an axe from scratch. We use numeric fluents for brevity, but the example does not require them.
\begin{itemize}
    \item $\mathcal{V} = \big\{
    N_{\textsc{food}}, 
    N_{\textsc{sticks}}, 
    N_{\textsc{stone}}
    \in \{0, 1,...,N\};\\
    \hphantom{\qquad\quad} \textsc{hungry}, \textsc{has\_axe} \in \{\textsc{true}, \textsc{false}\} \big\}$
    \item $\mathcal{O} = \{\\
    \texttt{get\_food}:\\
        \hphantom{\quad} \pre:\neg (N_{\textsc{food}} = N) \\
        \hphantom{\quad} \eff:(\mathrm{incr.}\ N_{\textsc{food}})\\
    \\
    \texttt{get\_stick}:\\
        \hphantom{\quad} \pre:\neg (N_{\textsc{sticks}} = N) \\
        \hphantom{\quad} \eff:(\mathrm{incr.}\ N_{\textsc{sticks}})\\
    \\
    \texttt{get\_stone}:\\
        \hphantom{\quad} \pre:\neg (N_{\textsc{stone}} = N) \\
        \hphantom{\quad} \eff:(\mathrm{incr.}\ N_{\textsc{stone}})\\
    \\
    \texttt{eat}:\\
        \hphantom{\quad} \pre:\neg (N_{\textsc{food}} = 0) \wedge \textsc{hungry} \\
        \hphantom{\quad} \eff:(\mathrm{decr.}\ N_{\textsc{food}} \wedge \neg \textsc{hungry} )\\
    \\
    \texttt{make\_axe}:\\
        \hphantom{\quad} \pre:\neg (N_{\textsc{sticks}} = 0 \vee N_{\textsc{stone}} = 0 \vee \textsc{has\_axe}) \\
        \hphantom{\quad} \eff:(\mathrm{decr.}\ N_{\textsc{sticks}} \wedge \mathrm{decr.}\ N_{\textsc{stone}} \wedge \textsc{has\_axe})\\
    \}$
    \item $s_0 = (0, 0, 0,  \textsc{false},  \textsc{false})$
    \item $G = (\neg \textsc{hungry} \wedge \textsc{has\_axe})$
    \\
\end{itemize}
All operators have unit cost. For this task, there are two optimal plans: [\texttt{get\_stick}, \texttt{get\_stone}, \texttt{make\_axe}], and [\texttt{get\_stone}, \texttt{get\_stick}, \texttt{make\_axe}]. The operators \texttt{get\_food} and \texttt{eat} are irrelevant and can be removed, since neither appears in any optimal plan.

\section{Task Scoping}
The purpose of task scoping is to identify and remove task-irrelevant variables and operators from the agent's model. This process produces an abstraction of the original planning problem aimed at making planning more tractable. However, not all abstractions preserve optimal plans, so the agent must carefully build its abstractions to avoid removing any relevant information.

\begin{definition}
Given a planning problem $P$, a \emph{task-scoping} abstraction $P'$ of $P$ is one that contains a subset of the variables and operators in $P$ such that all optimal plans in $P$ are still optimal plans of $P'$.
\end{definition}

In this section, we describe three types of task-scoping abstractions that remove increasing amounts of irrelevant information. We construct these abstractions using variations of Algorithm~\ref{alg:scoping}, and prove that each preserves optimal plans.

\subsection{Backwards Reachability of Variables}
\label{sec:backwards-reachability}
The first and simplest task-scoping abstraction encodes the notion that agents need not consider actions to be relevant unless they modify goal variables or preconditions of other relevant actions. This version of Algorithm~\ref{alg:scoping} (which we call Algorithm~\ref{alg:scoping}-a) omits any of the colored text appearing after the `$\|$' symbols.
It starts by considering only goal variables to be relevant, and performs backwards reachability analysis over variables, considering operators relevant if their effects contain any relevant variables, and then considering the precondition variables of any such operators to be relevant. The process repeats until no new variables are deemed relevant. 
The Fast Downward Planning System performs an equivalent abstraction during its knowledge compilation process \citep{Helmert:2006:FDP:1622559.1622565}.

In the example of Section~\ref{sec:open-scope-models}, this process would work proceed as follows:\vspace{6pt}\\
\setlength{\tabcolsep}{2pt}
\begin{tabular}{@{\hspace{10pt}}lllllll}
\textsc{has\_axe}& $\rightarrow$ &\texttt{make\_axe} &$\rightarrow$ &$N_{\textsc{sticks}}$ &$\rightarrow$ &\texttt{get\_stick}\\
                 &               &                   &$\rightarrow$ &$N_{\textsc{stone}}$ &$\rightarrow$ &\texttt{get\_stone}\\
\\
\textsc{hungry}  & $\rightarrow$ &\texttt{eat}       &$\rightarrow$ &$N_{\textsc{food}}$   &$\rightarrow$ &\texttt{get\_food}
\end{tabular}\vspace{6pt}
\noindent Eventually all variables and operators would be marked as relevant. Had $\neg\textsc{hungry}$ not appeared in the goal, the algorithm would not consider the bottom chain relevant, and those items would be removed. In Section~\ref{sec:causally-linked-irrelevance} we will upgrade the algorithm to recognize that $\neg\textsc{hungry}$ is satisfied by the initial state and not modified by the top operators; therefore, the bottom chain can still be removed.





\begin{algorithm}[t]
    \caption{\textsc{Task Scoping}}
    \label{alg:scoping}
    \textbf{Input}: $\langle \mathcal{V}, \mathcal{O}, s_0, G \rangle$\\
    \textbf{Output}: $\langle \mathcal{V}' \subseteq \mathcal{V}, \mathcal{K}' \subseteq \mathcal{V}, \mathcal{O}' \subseteq \mathcal{O}\rangle$
    \begin{algorithmic}[1] 
        \State $V_0 \gets \{ \textsc{dummy\_goal\_var}\}$ \Comment relevant vars
        \State $O_0 \gets \{\texttt{dummy\_goal\_operator}(G)\}$ \Comment relevant ops
        \Repeat
            \State $\overline{O}_i \gets O_{i-1}\ \|\ \textcolor{RedOrange}{\textsc{MergeSameEffects}(O_{i-1}, V_{i-1})}$
            \State $E_i \gets \varnothing\ \|\ \textcolor{RoyalBlue}{\{\text{variables} \in \eff(O_{i-1})\}}$
            \State $L_i \gets \varnothing\ \|\ \textcolor{RoyalBlue}{\textsc{CausalLinks}(s_0, E_i)}$
            \State $C_i \gets \{\text{clauses in $\pre(\overline{O}_i)$ 
            \textcolor{RoyalBlue}{not implied by $L_i$}}\}$
            \State $K_i \gets \{\text{vars in clauses of }\pre(\overline{O}_i)\} \setminus \vars(C_i)$ \\ \Comment causally linked variables
            \State $V_i \gets V_{i-1} \cup \{v \in \vars(c)\ \forall\ c \in C_i \}$
            \State $O_i \gets \{o \in \mathcal{O}: \eff(o) \cap V_i \neq \varnothing \} $
        \Until{$V_i = V_{i-1}$}
        \State $\mathcal{V}' \gets V_n \setminus \{\textsc{dummy\_goal\_var}\}$
        \State $\mathcal{K}' \gets K_n \setminus \{\textsc{dummy\_goal\_var}\}$
        \State $\mathcal{O}' \gets O_n \setminus \{\texttt{dummy\_goal\_operator}(G)\}$
        \State \Return $\langle \mathcal{V}', \mathcal{K}', \mathcal{O}'\rangle$
    \end{algorithmic}
\end{algorithm}

\begin{algorithm}[t]
    \caption{\textsc{MergeSameEffects}}
    \label{alg:merge-effects}
    \textbf{Input}: $O_i$, $V_i$\\
    \textbf{Output}: $\overline{O}_i$
    \begin{algorithmic}[1] 
        \State $O_{\mathrm{equiv}} \gets \text{Partition $O_i$ based on effects on $V_i$ and cost.}$
        \State $\overline{O}_i \gets \text{merge operators in equivalence classes: } \{$
        \Statex $\hphantom{\quad}\pre: \text{take disjunction of preconditions \& simplify}$
        \Statex $\hphantom{\quad}\eff: \text{copy effects on $V_i$ (identical)}$
        \Statex $\hphantom{\quad}\cost: \text{copy cost of component operators (identical)}$
        \Statex $\}$
        \State \Return $\overline{O}_i$
    \end{algorithmic}
\end{algorithm} 

\begin{algorithm}[t!]
    \caption{\textsc{CausalLinks}}
    \label{alg:causal-links}
    \textbf{Input}: $s_0$, $E_i$\\
    \textbf{Output}: $F_i$
    \begin{algorithmic}[1] 
        \State $F_i \gets \{\text{atoms in $s_0$ containing no vars in $E_i$}\}$
        \State \Return $F_i$
    \end{algorithmic}
\end{algorithm}

\subsection{Merging Same-Effect Operators}
\label{sec:merging-same-effect-ops}
The second task-scoping abstraction reflects the idea that actions which have identical effects on relevant variables are interchangeable and can therefore be merged. Algorithm~\ref{alg:scoping}-b extends the previous version by adding the \textsc{MergeSameEffects} function on line 4, which is detailed in Algorithm~\ref{alg:merge-effects}.

The merging procedure partitions the relevant operators $O_i$ into equivalence classes that have identical costs and effects on relevant variables $V_i$. Each resulting merged operator removes any effects on non-relevant variables, and its precondition is the disjunction of the original operators (simplified to remove any unnecessary clauses). As a result, line 7 of Algorithm~\ref{alg:scoping} now produces potentially fewer precondition clauses $C_i$ from which to add relevant variables in line 8. Note that the merged operators never appear in $\mathcal{O}'$, only the non-merged originals.

Suppose we modify the Minecraft domain of Section~\ref{sec:open-scope-models} to replace the \texttt{get\_food} operator with the following:

\begin{adjustwidth}{1em}{1em}
    $\texttt{hunt}:\\
        \hphantom{\quad} \pre:\neg (N_{\textsc{food}} = N) \wedge \neg \textsc{hungry} \\
        \hphantom{\quad} \eff:(\mathrm{incr.}\ N_{\textsc{food}}) \wedge \textsc{hungry}\\
        \\
    \texttt{gather}:\\
        \hphantom{\quad} \pre:\neg (N_{\textsc{food}} = N) \wedge \textsc{hungry}\\
        \hphantom{\quad} \eff:(\mathrm{incr.}\ N_{\textsc{food}})$
\end{adjustwidth}\vspace{3pt}
Furthermore, suppose $N_{\textsc{food}}$ is the only relevant variable (perhaps the task is now to gather food). Both operators modify $N_{\textsc{food}}$, so both would be marked as relevant. Algorithm~\ref{alg:scoping}\nobreakdash-b would then call \textsc{MergeSameEffects} and determine that both operators have the same effect on $N_{\textsc{food}}$ and can therefore be merged. After simplifying preconditions, the resulting merged operator would be equivalent to the original \texttt{get\_food} operator. In this example, \textsc{hungry} was not relevant to begin with and does not appear in the merged operator's precondition, so it would remain irrelevant. As a result, the \texttt{eat} operator, which modifies \textsc{hungry}, never becomes relevant either, and the algorithm will return $\mathcal{O}' = \{\texttt{hunt}, \texttt{gather}\}$.

\subsection{Causally Linked Irrelevance}
\label{sec:causally-linked-irrelevance}
The third task-scoping abstraction we introduce captures the idea that agents can additionally ignore variables that already match relevant preconditions and goal
clauses, unless a relevant action can modify them. This corresponds to the concept of causal links 
\citep{McAllester1991SystematicNP}.
A clause is \emph{causally linked} when (1) it is implied by some state (this work only considers causal links from $s_0$), (2) it appears in the precondition of a subsequent operator, and (3) it is not modified by any operators in between. In the Minecraft example of Section \ref{sec:open-scope-models}, this corresponded to the variable \textsc{hungry}. Since the initial state and goal both contained the clause $\neg \textsc{hungry}$ and none of the relevant operators modified it, \textsc{hungry} was \emph{causally linked} and could safely be removed.

The full version of Algorithm~\ref{alg:scoping} builds on the previous version by additionally identifying clauses $L_i$ (in lines 5-6) that are causally linked with the initial state $s_0$ and contain no overlap with any variables mentioned in the effects of relevant operators.\footnote{Additional optimizations are possible, such as by considering subsequent states or ignoring variables in $C_i$ when their values do not affect the truth value of the causally linked clauses.} Note that Algorithm~\ref{alg:scoping} is guaranteed to terminate, since $\mathcal{V}$ is finite and $V_{i-1} \subseteq V_i \subseteq \mathcal{V}$ for every iteration.

\subsection{Main Theorem}
We show in this section that Algorithm~\ref{alg:scoping} produces an abstraction of the original planning problem that contains all optimal plans. The proof works by removing unnecessary operators from each valid plan, and makes use of a Merge Substitution lemma, which ensures that the resulting plans are still valid.

For any quantity $X_i$ in the algorithm, we will use the subscript $X_n$ to indicate the value $X_i$ has in the final iteration of the algorithm. For example, $K_n$ is the final set of causally linked variables. Additionally, for a set of state variables $Z \subseteq \mathcal{V}$, we use the notation $s[Z]$ to denote the partial state of $s$ with respect to only the variables in $Z$.

\begin{lemma}[Merge Substitution]
\label{lemma:merge-substitution}

Let $o$ be any operator in $O_n$, $s$ any state that implies $\pre(o)$, and $s' \ne s$ another state. If the partial state of $s'$ with respect to relevant and causally linked variables is the same as in state $s$ (i.e. $s'[V_n \cup K_n] = s[V_n \cup K_n]$), then there exists another operator $o'$, also in $O_n$ (and possibly equal to $o$), such that: $s'$ implies $\pre(o')$; $o$ and $o'$ have the same effect on $V_n$; neither $o$ nor $o'$ has any effect on $K_n$; and $o$ and $o'$ have the same cost.

\begin{proof}
Run $\textsc{MergeSameEffects}(O_n, V_n)$ to compute $\overline{O}_n$, and find the (potentially merged) operator $\overline{o}$ corresponding to $o$. Such an operator exists, because \textsc{MergeSameEffects} partitions $O_n$.

First we will show that $s'$ implies $\pre(\overline{o})$.
By construction, $\pre(\overline{o})$ contains only variables in $(V_n \cup K_n)$, and the clauses containing variables in $K_n$ are causally linked. This means that if $s$ implies $\pre(\overline{o})$, and $s'[V_n \cup K_n] = s[V_n \cup K_n]$, then $s'$ also implies $\pre(\overline{o})$.
Since $\pre(\overline{o})$ is just the disjunction of the preconditions of its component operators, it must be the case that $s'$ implies the precondition of at least one such component operator $o' \in O_n$.

Since $o'$ and $o$ correspond to the same abstract operator $\overline{o}$, they must have the same effect on $V_n$ and the same cost. Since both operators are in $O_n$, neither can affect $K_n$.
\end{proof}
\end{lemma}

\begin{theorem}
Given a planning problem, all optimal plans use a subset of the operators returned by Algorithm~\ref{alg:scoping}.

\begin{proof}

We will show that for any valid plan $\pi$ containing operators not in $O_n$, there exists another plan $\pi'$ that is also valid, shorter, has less cost, and only uses operators in $O_n$.

We will go through the operators of $\pi$, from the beginning to the end, and for each operator $o$, add a corresponding operator to $\pi'$ as follows:

\begin{enumerate}
    \item If $o$ affects at least one variable in $V_n$ and can be taken from the current state of the modified plan, keep it; $o$ is in $O_n$, since it modifies a variable in $V_n$.
    \item If $o$ affects at least one variable in $V_n$ and cannot be taken from the current state of the modified plan, replace it with an operator $o'$ that has the same effects on $V_n$ and can be taken from the current state.
    \item If $o$ does not affect $V_n$ (and therefore is not in $O_n$), add a no-op operator to $\pi'$ with the same cost as $o$.
\end{enumerate}

Note that in case 3, the no-op operators need not exist in the domain. We could simply ignore $o$ entirely in this case, but this would make the correspondences between the operators and states of $\pi$ and $\pi'$ slightly less clear, since the plans would have different lengths. We will delete the no-ops from $\pi'$ after comparing the plans.

Now we provide an inductive proof over the steps in the plan to show that the following inductive assumption holds.

\textbf{Inductive assumption:}  At each step, $\pi$ and $\pi'$ have the same partial state on $V_n$, and $\pi'$'s partial state on $K_n$ (the set of causally linked variables) is equal to the initial partial state on $K_n$. That is, $s_i[V_n] = s'_i[V_n]$ and $s'_i[K_n] = s_0[K_n]$.

\textbf{Inductive base:} Empty plans share initial state.

\textbf{Inductive step:} Each of the three cases outlined above preserves the inductive assumption.
\begin{itemize}
    \item Whenever case 1 applies, the original operator $o$ is applicable and does not change the plan. The resulting partial state $s'_{i+1}[V_n] = s_{i+1}[V_n]$. Since $o$ is in $O_n$, it does not modify $K_n$.
    \item Whenever case 2 applies, this operator replacement is possible due to Lemma~\ref{lemma:merge-substitution} and the inductive assumption. By Lemma~\ref{lemma:merge-substitution}, $o$ and $o'$ have the same effect on $V_n$, no effect on $K_n$, and equal cost. Furthermore, $o'$ is in $O_n$.
    \item Whenever case 3 applies, $o$ does not modify $V_n$. Therefore $s'_{i+1}[V_n] = s_{i+1}[V_n]$. The no-op does not affect $K_n$, and has equal cost to $o$.
\end{itemize}
The above inductive argument shows that $\pi'$ is also a valid plan, and has the same cost and length as $\pi$.

We will now delete the no-ops from $\pi'$. Afterwards, $\pi'$ is shorter and has lower cost than $\pi$, and uses only operators from $O_n$.

Since for any valid plan $\pi$ containing an operator outside of $O_n$ we can produce a shorter and cheaper plan, it must be the case that all optimal plans use only operators from $O_n$.
\end{proof}
\end{theorem}

\subsection{Discussion}
\label{sec:discussion}

It should be clear that Algorithm~\ref{alg:scoping} does not involve searching through specific states. Rather, it reasons over the variables and operators of the planning problem. Space precludes a thorough complexity analysis, but Algorithm~\ref{alg:scoping}'s worst-case complexity is dominated by $|\mathcal{V}| \times |\mathcal{O}|$. Since $|\mathcal{V}| \times |\mathcal{O}|$ is generally much smaller than the problem's full state--action space (which may even be infinite), building a task-scoping abstraction is thus often significantly more efficient than planning over the original problem.\footnote{This complexity expression neglects the cost of simplifying preconditions in Algorithm~\ref{alg:scoping}, which is reasonable in practice because most preconditions are relatively simple.}

The abstractions in the previous sections are not intended to be an exhaustive list of task-scoping abstractions. More aggressive abstractions are clearly possible, particularly when considering causal links.

\section{Experimental Evaluation}
\label{sec:experiments}
Algorithm~\ref{alg:scoping} is agnostic to which particular representation is used to express the planning problem. To demonstrate its performance and utility empirically, we customized it to work with both numeric PDDL 2.1 Level 2, and SAS+ \citep{Helmert:2006:FDP:1622559.1622565}.
The output of the algorithm  consists of relevant variables $\mathcal{V}'$, causally linked variables $\mathcal{K}'$, and relevant operators $\mathcal{O}'$.
For SAS+, we simply remove operators outside of $\mathcal{O}'$ and causally linked goal conditions.
For PDDL, we remove lifted operators from the domain file whenever they correspond to no grounded operators in $\mathcal{O}'$, and we remove objects from the problem file whenever they correspond to no variables in $\mathcal{V}' \cup \mathcal{K}'$.\footnote{
We keep objects corresponding to variables in $\mathcal{K}'$ because these objects may be used to ground operators in $\mathcal{O}'$.
}
Since this is a more conservative abstraction, it is still a valid task-scoping abstraction.

All experiments were conducted on a cluster of 2.90GHz Intel Xeon Platinum 8268 CPUs, using 2 virtual cores and 16GB of RAM per trial, and measurements are averaged across 10 independent trials.

\begin{figure}[t]
\includegraphics[width=\linewidth]{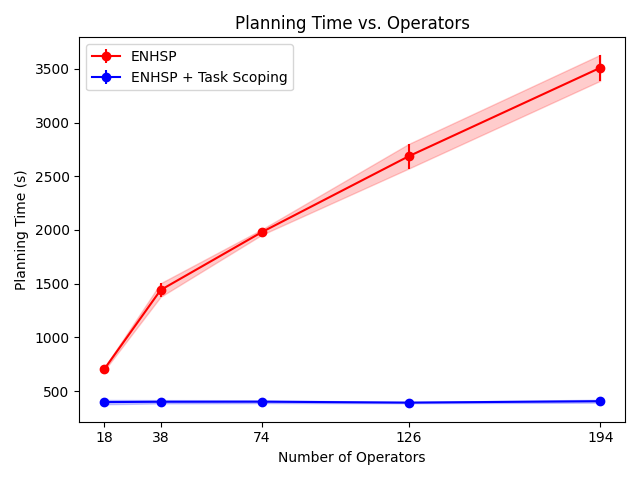}
\caption{\small{Results for the Multi-Switch Continuous Playroom domain. Total planning time with task scoping is essentially constant compared to the baseline, even as the size of the domain grows exponentially. (Total planning time includes scoping time; error bars show standard deviation across 10 independent trials.)}}
\vspace{-0.25em}
\label{fig:MonkeyResults}
\end{figure}

\begin{table*}[t]
\resizebox{\textwidth}{!}{%
\begin{tabular}{lrrrrrrrrrr}
\toprule
\multicolumn{1}{l}{\textbf{Problem}} & \multicolumn{2}{c}{\textbf{Operators}}                             & \multicolumn{2}{c}{\textbf{Expansions}}                            & \multicolumn{2}{c}{\textbf{Evaluations}}                           & \multicolumn{1}{c}{\textbf{Scoping}} & \multicolumn{1}{c}{\textbf{Planning}} & \multicolumn{2}{c}{\textbf{Total Time (s)}}  \\
\multicolumn{1}{c}{}        & \multicolumn{1}{c}{Unscoped} & \multicolumn{1}{c}{Scoped} & \multicolumn{1}{c}{Unscoped} & \multicolumn{1}{c}{Scoped} & \multicolumn{1}{c}{Unscoped} & \multicolumn{1}{c}{Scoped} & \multicolumn{1}{c}{}            & \multicolumn{1}{c}{Scoped}                       & \multicolumn{1}{c}{Unscoped}    & \multicolumn{1}{c}{Scoped} \\
\midrule
Playroom 1 & 18 & \textbf{14} & 8,635,588 & \textbf{7,658,632} & 13,020,860 & \textbf{11,294,750} & $0.3 \pm 0.1$ & $397.0 \pm 20.1$ & $702.9 \pm 13.1$ & $\mathbf{397.3 \pm 20.1}$ \\
Playroom 3 & 38 & \textbf{14} & 9,589,906 & \textbf{7,658,632} & 14,930,660 & \textbf{11,294,750} & $0.5 \pm 0.0$ & $400.0 \pm 15.4$ & $1440.4 \pm 63.5$ & $\mathbf{400.6 \pm 15.4}$ \\
Playroom 5 & 74 & \textbf{14} & 9,685,772 & \textbf{7,658,632} & 15,217,470 & \textbf{11,294,750} & $0.9 \pm 0.0$ & $400.2 \pm 13.2$ & $1981.0 \pm 25.2$ & $\mathbf{401.1 \pm 13.3}$ \\
Playroom 7 & 126 & \textbf{14} & 9,685,872 & \textbf{7,658,632} & 15,218,910 & \textbf{11,294,750} & $1.5 \pm 0.1$ & $391.0 \pm 5.9$ & $2685.7 \pm 114.6$ & $\mathbf{392.5 \pm 5.9}$ \\
Playroom 9 & 194 & \textbf{14} & 9,685,872 & \textbf{7,658,632} & 15,218,910 & \textbf{11,294,750} & $2.3 \pm 0.1$ & $403.3 \pm 13.7$ & $3509.8 \pm 121.1$ & $\mathbf{405.6 \pm 13.7}$ \\
\midrule
Composite (depot) & 1809 & \textbf{1062} & $(> 2.7\mathrm{M})$ & \textbf{184,218} & $(>19.0\mathrm{M})$ & \textbf{487,858} & $26.6 \pm 0.3$ & $84.9 \pm 0.9$ & $(>5.8\mathrm{K})$ & $\mathbf{111.5 \pm 1.1}$ \\
Composite (driverlog) & 1809 & \textbf{336} & $(> 2.9\mathrm{M})$ & \textbf{54,282} & $(> 18.6\mathrm{M})$ & \textbf{66,332} & $22.6 \pm 0.1$ & $4.7 \pm 0.3$ & $(> 4.3\mathrm{K}) $ & $\mathbf{27.3 \pm 0.4}$ \\
Composite (satellite) & 1809 & \textbf{94} & $(> 2.5\mathrm{M})$ & \textbf{16,916} & $(> 22.1\mathrm{M})$ & \textbf{91,805} & $23.4 \pm 0.2$ & $3.8 \pm 0.4$ & $(> 5.3\mathrm{K})$ & $\mathbf{27.2 \pm 0.4}$ \\
Composite (zenotravel) & 1809 & \textbf{37} & 283,797 & \textbf{191} & 1,729,374 & \textbf{598} & $11.7 \pm 0.1$ & $1.1 \pm 0.1$ & $362.6 \pm 1.8$ & $\mathbf{12.7 \pm 0.1}$ \\
\midrule
Minecraft (planks) & 106 & \textbf{22} & 64 & 64 & 353 & \textbf{338} & $4.8 \pm 0.1$ & $1.3 \pm 0.2$ & $\mathbf{1.7 \pm 0.2}$ & $6.1 \pm 0.3$ \\
Minecraft (wool) & 106 & \textbf{18} & 220,959 & \textbf{2,736} & 1,615,714 & \textbf{6,612} & $4.1 \pm 0.1$ & $1.7 \pm 0.2$ & $432.5 \pm 21.5$ & $\mathbf{5.8 \pm 0.2}$ \\
Minecraft (bed) & 106 & \textbf{25} & $(>2.3\mathrm{M})$ & \textbf{182,092} & $(> 14.2\mathrm{M})$ & \textbf{1,136,831} & $5.0 \pm 0.1$ & $173.1 \pm 8.0$ & $(>4.3\mathrm{K})$ & $\mathbf{178.1 \pm 8.0}$ \\
\bottomrule
\end{tabular}
}
\caption{\small{Numeric planning results. Task scoping removes irrelevant operators, which leads to significant reductions in node expansions, evaluations, and planning time. Boldface denotes better performance. Standard deviation is across $10$ trials. $(> N)$ denotes out-of-memory.}}
\label{table:experiment_times_enhsp}
\end{table*}

\subsection{Numeric Domains with ENHSP-2020}
\label{expers:ENHSP_Experiments}
We first investigate our approach's performance and utility in numeric domains. In all the following experiments, we run ENHSP-2020 \citep{Scala2020SubgoalingTF} in optimal mode (\texttt{WAStar+hrmax}), both with and without task scoping as a pre-processing step. We measure the end-to-end wall-clock time, as well as nodes generated during search, to produce a plan.

\paragraph{Multi-Switch Continuous Playroom.}
In order to study how our approach scales with the size of the input problem in a simplified setting, we implemented the continuous playroom domain from \citet{playroompaper} in PDDL 2.1.
In this domain, the agent can move in cardinal directions within a grid. Its goal is to pick up a ball and throw it at a bell, but this requires first pressing a number of green buttons, which in turn requires turning on a number of lights. Both the lights and buttons are strewn throughout the grid.
In our experiments, we create progressively larger problems with more buttons and lights, yet ensure that all green buttons are turned on in the initial state. This renders the lights causally-linked with the initial state, so all corresponding actions that affect them are irrelevant.

The results, shown in Figure~\ref{fig:MonkeyResults}, demonstrate that task scoping is able to keep planning time relatively constant compared to the baseline, even as the size of the problem grows exponentially.
Table~\ref{table:experiment_times_enhsp} provides more detail, and shows that even after accounting for the time required to perform task scoping, the scoped planner still finds plans more quickly than the unscoped baseline.


\paragraph{Composite IPC Domains}
Task scoping is intended to make \emph{open-scope} planning tractable, but most existing benchmarks are not open-scope: they have been carefully hand-designed to exclude any task-irrelevant information. While a full investigation of how to \emph{learn} such an open-scope model is beyond the scope of this paper, it is straightforward to construct an open-scope domain from existing benchmarks: simply combine multiple domains together and attempt to solve a problem from any one of them.

We construct such a model by combining the domain and problem files of the \texttt{Depots}, \texttt{DriverLog}, \texttt{Satellite}, and \texttt{ZenoTravel} domains from IPC 2002 \citep{IPC_2002} into a single composite domain file and 4 different composite problem files. The problem files differ only in their goal condition: each file contains a goal related to a specific sub-domain, and the variables and actions related to the other 3 domains are task-irrelevant. 

The results in Table~\ref{table:experiment_times_enhsp} indicate that task scoping can dramatically reduce the number of operators, which makes optimal open-scope planning tractable in all four tasks, whereas ENHSP typically runs out of memory without scoping.

\paragraph{Minecraft}
To examine the utility of task scoping on a novel open-scope domain of interest, we created a planning domain containing simplified dynamics and several tasks from Minecraft and pictured in Figure~\ref{fig:FrontPage}. The domain features interactive items at various locations in the map, which the agent can destroy, place elsewhere, or use to craft different items. Thus, the domain is truly open-scope: it supports a large variety of potential tasks such that most objects and actions are irrelevant to most tasks. Within this domain, we wrote PDDL 2.1 files to express 3 specific tasks: (1) craft wooden planks, (2) dye 3 wool blocks blue, and (3) craft and place a blue bed at a specific location (which requires completing both prior tasks as subgoals). The results show that task scoping is able to recognize and remove a large number of irrelevant operators depending on the task chosen within this domain, as shown in Table~\ref{table:experiment_times_enhsp}.
This dramatically speeds up planning time for the wool-dyeing and bed-making tasks.\footnote{We also created a propositional version of this domain. Fast Downward was not even able to complete translation on it; however, after removing irrelevant objects for each problem by hand, planning took just a few seconds.}

\begin{table*}[t]
\resizebox{\textwidth}{!}{%
\begin{tabular}{lrrrrrrrrrrrrr}
\toprule
\textbf{Problem} & \multicolumn{2}{c}{\textbf{Operators}} & \multicolumn{2}{c}{\textbf{Expansions}} & \multicolumn{2}{c}{\textbf{Evaluations}} & \textbf{Translate} & \textbf{Scoping} & \multicolumn{2}{c}{\textbf{Planning Time (s)}} & \multicolumn{2}{c}{\textbf{Total Time (s)}}\\
  & Unscoped & \multicolumn{1}{c}{Scoped} & Unscoped & \multicolumn{1}{c}{Scoped} & Unscoped & \multicolumn{1}{c}{Scoped} &  &  & Unscoped & \multicolumn{1}{c}{Scoped} & Unscoped & \multicolumn{1}{c}{Scoped}\\
\midrule
Driverlog 15 & $2592$ & $\mathbf{2112}$ & $1,392$ & $\mathbf{1,379}$ & $22,980$ & $\mathbf{21,186}$ & $0.5 \pm 0.0$ & $3.6 \pm 0.2$ & $4.4 \pm 0.1$ & $\mathbf{3.1 \pm 0.1}$ & $\mathbf{4.9 \pm 0.2}$ & $7.2 \pm 0.2$\\
Driverlog 16 & $4890$ & $\mathbf{3540}$ & $3,618$ & $\mathbf{3,087}$ & $87,465$ & $\mathbf{60,306}$ & $0.7 \pm 0.0$ & $8.3 \pm 0.3$ & $19.8 \pm 0.8$ & $\mathbf{8.1 \pm 0.2}$ & $20.5 \pm 0.8$ & $\mathbf{17.1 \pm 0.5}$\\
Driverlog 17 & $6170$ & $\mathbf{3770}$ & $1,058$ & $\mathbf{985}$ & $28,926$ & $\mathbf{21,030}$ & $0.8 \pm 0.0$ & $9.6 \pm 0.3$ & $22.0 \pm 0.9$ & $\mathbf{7.9 \pm 0.2}$ & $22.8 \pm 1.0$ & $\mathbf{18.3 \pm 0.5}$\\
Logistics 15 & $650$ & $\mathbf{250}$ & $6,395$ & $6,395$ & $153,488$ & $\mathbf{117,643}$ & $0.3 \pm 0.1$ & $0.7 \pm 0.1$ & $11.6 \pm 0.2$ & $\mathbf{3.3 \pm 0.1}$ & $11.9 \pm 0.2$ & $\mathbf{4.2 \pm 0.1}$\\
Logistics 20 & $650$ & $\mathbf{250}$ & $14,798$ & $\mathbf{14,390}$ & $381,235$ & $\mathbf{260,336}$ & $0.3 \pm 0.0$ & $0.7 \pm 0.0$ & $26.9 \pm 0.3$ & $\mathbf{5.9 \pm 0.1}$ & $27.2 \pm 0.3$ & $\mathbf{6.9 \pm 0.2}$\\
Logistics 25 & $650$ & $\mathbf{290}$ & $67,931$ & $\mathbf{66,683}$ & $1,701,778$ & $\mathbf{1,281,693}$ & $0.3 \pm 0.0$ & $0.8 \pm 0.0$ & $127.7 \pm 1.4$ & $\mathbf{34.8 \pm 0.4}$ & $128.0 \pm 1.4$ & $\mathbf{35.8 \pm 0.4}$\\
Satellite 05 & $609$ & $\mathbf{339}$ & $1,034$ & $1,034$ & $63,122$ & $\mathbf{35,231}$ & $0.3 \pm 0.1$ & $0.6 \pm 0.1$ & $2.0 \pm 0.0$ & $\mathbf{0.6 \pm 0.0}$ & $2.3 \pm 0.1$ & $\mathbf{1.5 \pm 0.0}$\\
Satellite 06 & $582$ & $\mathbf{362}$ & $5,766$ & $\mathbf{4,886}$ & $311,695$ & $\mathbf{166,415}$ & $0.3 \pm 0.1$ & $0.5 \pm 0.1$ & $6.3 \pm 0.1$ & $\mathbf{1.7 \pm 0.0}$ & $6.6 \pm 0.1$ & $\mathbf{2.5 \pm 0.0}$\\
Satellite 07 & $983$ & $\mathbf{587}$ & $124,703$ & $\mathbf{96,125}$ & $10,498,090$ & $\mathbf{4,915,020}$ & $0.4 \pm 0.1$ & $0.9 \pm 0.1$ & $333.1 \pm 1.8$ & $\mathbf{50.9 \pm 0.3}$ & $333.4 \pm 1.8$ & $\mathbf{52.2 \pm 0.3}$\\
Zenotravel 10 & $1155$ & $\mathbf{1095}$ & $23,661$ & $\mathbf{23,649}$ & $675,945$ & $\mathbf{654,968}$ & $0.3 \pm 0.0$ & $2.4 \pm 0.1$ & $37.7 \pm 0.7$ & $\mathbf{33.2 \pm 0.4}$ & $38.1 \pm 0.7$ & $\mathbf{35.9 \pm 0.5}$\\
Zenotravel 12 & $3375$ & $\mathbf{3159}$ & $4,766$ & $\mathbf{4,735}$ & $222,157$ & $\mathbf{211,183}$ & $0.6 \pm 0.0$ & $7.4 \pm 0.1$ & $45.5 \pm 0.1$ & $\mathbf{39.2 \pm 0.3}$ & $\mathbf{46.0 \pm 0.2}$ & $47.2 \pm 0.3$\\
Zenotravel 14 & $6700$ & $\mathbf{6200}$ & $6,539$ & $6,539$ & $598,578$ & $\mathbf{587,771}$ & $1.0 \pm 0.0$ & $14.2 \pm 0.2$ & $232.4 \pm 18.5$ & $\mathbf{193.2 \pm 8.7}$ & $233.4 \pm 18.5$ & $\mathbf{208.3 \pm 8.6}$\\
\bottomrule
\end{tabular}
}
\caption{\small{Classical planning results. Task scoping removes irrelevant operators from every domain, which leads to significant reductions in node expansions, evaluations, and planning time. Boldface denotes better performance. Standard deviation is across 10 trials.}}
\label{table:experiment_times_fd}
\end{table*}


\subsection{Classical Planning with Fast Downward}
\label{expers:FD_Experiments}
Having investigated our approach's utility and performance in a variety of numeric domains, we now turn to propositional domains.
We are interested in examining whether our approach is able to discover task-irrelevant information beyond what the translator component of the well-known Fast-Downward planning system \citep{Helmert:2006:FDP:1622559.1622565}, and whether removing such irrelevance can substantially improve planning time.
To this end, we selected 4 benchmark domains (Logistics, DriverLog, Satellite, Zenotravel) from the optimal track of several previous iterations of the International Planning Competition (IPC) \citep{IPC_2002,IPC2014,IPC2009,IPC2000}. Since these domains do not contain any task-irrelevance on their own \citep{AIPlanningPerspectiveOnAbstraction}, 
we modified 3 problem files from each domain (Logistics, DriverLog, Satellite and Zenotravel) with initial states and goals set to introduce irrelevance while keeping the domain files fixed.
We translated each problem to SAS+ using FD's translator, ran task scoping on the resulting SAS+ file, then ran the FD planner with the LM-cut heuristic \citep{lmcut} on this problem. 
We report number of operators in the problem, as well as the time taken and nodes expanded and evaluated during search both with and without task scoping.

The results (see Table \ref{table:experiment_times_fd}) reveal that task scoping can abstract some of these problems beyond what FD's translator can accomplish alone and lead to a net speedup. Algorithm~\ref{alg:scoping} reduces the number of operators significantly for all $4$ domains. This difference was mostly because FD's translator was unable to remove any causally-linked irrelevant variables or operators, though it was able to remove the simpler types of irrelevance discussed in Section \ref{sec:backwards-reachability}.

\section{Related Work}
\label{sec:relatedwork}

The Fast Downward Planning System \citep{Helmert:2006:FDP:1622559.1622565} performs Algorithm~\ref{alg:scoping}-a from Section \ref{sec:backwards-reachability} as part of its knowledge compilation process.
This is backwards reachability analysis on what the authors call the \textit{achievability} causal graph.
The \textsc{MergeSameEffects} extension can be interpreted as computing abstract operators with fewer preconditions, making the causal graph sparser.
The \textsc{CausalLinks} extension also makes the causal graph sparser by ignoring satisfied clauses of preconditions.
The sparser causal graph means that the backwards reachability analysis terminates sooner, with fewer variables marked as potentially relevant.

Another popular method for reducing the size of the search space is the discovery of forward and backward invariants (a.k.a mutex constraints) \citep{BonetAndGeffnerOrigMutexPaper,EdelkampAndHelmertMinimizing,ChenLongMutexes,AlcazarReminderPaper}. Removing such invariants removes unreachable states or dead-end states, and their associated operators, from the planning problem and has been shown to dramatically improve search \citep{Helmert:2006:FDP:1622559.1622565}. However, removing invariants essentially amounts to removing states and operators that cannot be part of any valid plan. By contrast, our approach removes states that are very much reachable from both the initial and goal states; removing them does not preserve all valid plans, but rather all \textit{optimal} plans.

Some recent work removes operators and corresponding states \citep{FivserOpMutexes,horcik2021endomorphisms} that may be part of valid plans, but can still be safely ignored to preserve at least one optimal plan. This research is based on the central idea that some operators, or transitions~\citep{haslumSafeStrongRelevance,torralbaFocusing}, in valid plans may be strictly dominated by others, and thus can be safely removed. Our work can be seen as focusing on efficiently removing a subset of such dominated operators and states. Importantly, these existing approaches depend on problems having a finite state space---they often rely on ``factorizing" a problem into smaller problems~\citep{horcik2021endomorphisms,torralba2015simulation} and performing potentially expensive operations like symmetry-checking or constraint-satisfaction over these smaller problems. By contrast, our approach can handle infinite state spaces (as long as the number of variables and operators is finite), and Algorithm~\ref{alg:scoping}'s complexity does not scale with the size of the state space, but rather with the number of (grounded) variables and operators. 

Yet another line of work involves using abstractions to derive heuristics to guide search within the concrete problem \citep{PatternDatabases,NebelRelevanceHeuristic,KatzImplicitAbs}. Some of these approaches can use richer families of abstractions than Algorithm~\ref{alg:scoping} (for example, Cartesian abstractions). However, such approaches do not directly remove irrelevance from planning tasks, since the resulting abstractions do not necessarily preserve any valid plans. Some of this research~\citep{cegar_pattern,cegarjair} performs an iterative abstraction refinement similar to our approach, but interleaves planning and abstraction refinement whereas Algorithm~\ref{alg:scoping} does not need to perform planning to refine its abstraction. 



\section{Conclusion}
Task scoping enables existing domain-independent planners to generalize to a much broader class of \emph{open-scope} planning problems.
By carefully removing irrelevant variables and actions from consideration, our algorithm allows planners to overcome the exponential cost of planning with large amounts of irrelevant information.
This reduction in problem complexity leads to substantial improvements in planning time, even after accounting for the time spent building such abstractions.
Moreover, planners that use these abstractions do not suffer any penalty in terms of plan quality, as all optimal plans are guaranteed to be preserved under our algorithm.
This work builds on the already impressive legacy of domain-independent planners as general-purpose problem solvers, and represents an important step on the path to realizing truly general decision-making agents.

\bibliographystyle{named}
\bibliography{references}

\begin{thebibliography}{}

\bibitem[\protect\citeauthoryear{Alc\'{a}zar and
  Torralba}{2015}]{AlcazarReminderPaper}
Vidal Alc\'{a}zar and \'{A}lvaro Torralba.
\newblock A reminder about the importance of computing and exploiting
  invariants in planning.
\newblock In {\em ICAPS}, page 2–6, 2015.

\bibitem[\protect\citeauthoryear{Bonet and
  Geffner}{2001}]{BonetAndGeffnerOrigMutexPaper}
Blai Bonet and Héctor Geffner.
\newblock Planning as heuristic search.
\newblock {\em Artificial Intelligence}, 129(1):5--33, 2001.

\bibitem[\protect\citeauthoryear{Chen \bgroup \em et al.\egroup
  }{2007}]{ChenLongMutexes}
Yixin Chen, Zhao Xing, and Weixiong Zhang.
\newblock Long-distance mutual exclusion for propositional planning.
\newblock In {\em IJCAI}, page 1840–1845, 2007.

\bibitem[\protect\citeauthoryear{Chentanez \bgroup \em et al.\egroup
  }{2005}]{playroompaper}
Nuttapong Chentanez, Andrew~G Barto, and Satinder~P Singh.
\newblock Intrinsically motivated reinforcement learning.
\newblock In {\em NIPS}, pages 1281--1288, 2005.

\bibitem[\protect\citeauthoryear{Culberson and
  Schaeffer}{1998}]{PatternDatabases}
Joseph~C. Culberson and Jonathan Schaeffer.
\newblock Pattern databases.
\newblock {\em Computational Intelligence}, 14(3):318--334, 1998.

\bibitem[\protect\citeauthoryear{Edelkamp and
  Helmert}{1999}]{EdelkampAndHelmertMinimizing}
Stefan Edelkamp and Malte Helmert.
\newblock Exhibiting knowledge in planning problems to minimize state encoding
  length.
\newblock In {\em ECP}, page 135–147, 1999.

\bibitem[\protect\citeauthoryear{Fi{\v{s}}er \bgroup \em et al.\egroup
  }{2019}]{FivserOpMutexes}
Daniel Fi{\v{s}}er, {\'{A}}lvaro Torralba, and Alexander Shleyfman.
\newblock Operator mutexes and symmetries for simplifying planning tasks.
\newblock In {\em AAAI}, pages 7586--7593, 2019.

\bibitem[\protect\citeauthoryear{Fox and Long}{2003}]{PDDL2.1}
Maria Fox and Derek Long.
\newblock Pddl2. 1: An extension to pddl for expressing temporal planning
  domains.
\newblock {\em Journal of artificial intelligence research}, 20:61--124, 2003.

\bibitem[\protect\citeauthoryear{Gerevini \bgroup \em et al.\egroup
  }{2009}]{IPC2009}
Alfonso~E Gerevini, Patrik Haslum, Derek Long, Alessandro Saetti, and Yannis
  Dimopoulos.
\newblock Deterministic planning in the fifth international planning
  competition: Pddl3 and experimental evaluation of the planners.
\newblock {\em Artificial Intelligence}, 173(5-6):619--668, 2009.

\bibitem[\protect\citeauthoryear{Haslum \bgroup \em et al.\egroup
  }{2013}]{haslumSafeStrongRelevance}
Patrik Haslum, Malte Helmert, and Anders Jonsson.
\newblock Safe, strong, and tractable relevance analysis for planning.
\newblock In {\em ICAPS}, pages 317--321, 2013.

\bibitem[\protect\citeauthoryear{Helmert and Domshlak}{2009}]{lmcut}
Malte Helmert and Carmel Domshlak.
\newblock Landmarks, critical paths and abstractions: What's the difference
  anyway?
\newblock In {\em ICAPS}, pages 162--169, 2009.

\bibitem[\protect\citeauthoryear{Helmert}{2006}]{Helmert:2006:FDP:1622559.1622565}
Malte Helmert.
\newblock The fast downward planning system.
\newblock {\em J. Artif. Int. Res.}, 26(1):191--246, 2006.

\bibitem[\protect\citeauthoryear{Hoffmann \bgroup \em et al.\egroup
  }{2006}]{AIPlanningPerspectiveOnAbstraction}
Jörg Hoffmann, Ashish Sabharwal, Carmel Domshlak, Derek Long, Stephen Smith,
  Daniel Borrajo, and Lee McCluskey.
\newblock Friends or foes? an ai planning perspective on abstraction and
  search.
\newblock In {\em ICAPS}, pages 294--303, 2006.

\bibitem[\protect\citeauthoryear{Horčík and
  Fišer}{2021}]{horcik2021endomorphisms}
Rostislav Horčík and Daniel Fišer.
\newblock Endomorphisms of classical planning tasks.
\newblock In {\em AAAI}, pages 11835--11843, 2021.

\bibitem[\protect\citeauthoryear{Katz and Domshlak}{2010}]{KatzImplicitAbs}
Michael Katz and Carmel Domshlak.
\newblock Implicit abstraction heuristics.
\newblock {\em J. Artif. Intell. Res.}, 39:51--126, 2010.

\bibitem[\protect\citeauthoryear{Konidaris}{2019}]{GeorgeNecessity}
George Konidaris.
\newblock On the necessity of abstraction.
\newblock {\em Current Opinion in Behavioral Sciences}, 29:1 -- 7, 2019.
\newblock SI: 29: Artificial Intelligence (2019).

\bibitem[\protect\citeauthoryear{Korf}{1985a}]{KORF198597}
Richard~E. Korf.
\newblock Depth-first iterative-deepening: An optimal admissible tree search.
\newblock {\em Artificial Intelligence}, 27(1):97--109, 1985.

\bibitem[\protect\citeauthoryear{Korf}{1985b}]{KORF198535}
Richard~E. Korf.
\newblock Macro-operators: A weak method for learning.
\newblock {\em Artificial Intelligence}, 26(1):35--77, 1985.

\bibitem[\protect\citeauthoryear{Long and Fox}{2003}]{IPC_2002}
Derek Long and Maria Fox.
\newblock The 3rd international planning competition: Results and analysis.
\newblock {\em J. Artif. Intell. Res.}, 20:1--59, 2003.

\bibitem[\protect\citeauthoryear{Long \bgroup \em et al.\egroup
  }{2000}]{IPC2000}
Derek Long, Henry Kautz, Bart Selman, Blai Bonet, Hector Geffner, Jana Koehler,
  Michael Brenner, Joerg Hoffmann, Frank Rittinger, Corin~R Anderson, et~al.
\newblock The aips-98 planning competition.
\newblock {\em AI magazine}, 21(2):13--13, 2000.

\bibitem[\protect\citeauthoryear{Matloob and
  Soutchanski}{2016}]{Matloob2016ExploringOS}
Rami Matloob and Mikhail Soutchanski.
\newblock Exploring organic synthesis with state-of-the-art planning
  techniques.
\newblock In {\em Proceedings of Scheduling and Planning Applications woRKshop
  (SPARK)}, 2016.

\bibitem[\protect\citeauthoryear{McAllester and
  Rosenblitt}{1991}]{McAllester1991SystematicNP}
David~A. McAllester and David Rosenblitt.
\newblock Systematic nonlinear planning.
\newblock In {\em AAAI Conference on Artificial Intelligence}, 1991.

\bibitem[\protect\citeauthoryear{Nebel \bgroup \em et al.\egroup
  }{1997}]{NebelRelevanceHeuristic}
Bernhard Nebel, Yannis Dimopoulos, and Jana Koehler.
\newblock Ignoring irrelevant facts and operators in plan generation.
\newblock In {\em ECP}, pages 338--350, 1997.

\bibitem[\protect\citeauthoryear{Refanidis \bgroup \em et al.\egroup
  }{2001}]{planning_for_transportation}
Ioannis Refanidis, Nick Bassiliades, I.~Vlahavas, and Thessaloniki Greece.
\newblock Ai planning for transportation logistics.
\newblock {\em Proceedings 17th International Logistics Conference}, 12 2001.

\bibitem[\protect\citeauthoryear{Rovner \bgroup \em et al.\egroup
  }{2019}]{cegar_pattern}
Alexander Rovner, Silvan Sievers, and Malte Helmert.
\newblock Counterexample-guided abstraction refinement for pattern selection in
  optimal classical planning.
\newblock In {\em ICAPS}, pages 362--367, 2019.

\bibitem[\protect\citeauthoryear{Sandewall and
  R{\"o}nnquist}{1986}]{Sandewall1986ARO}
Erik Sandewall and Ralph R{\"o}nnquist.
\newblock A representation of action structures.
\newblock In {\em AAAI}, 1986.

\bibitem[\protect\citeauthoryear{Scala \bgroup \em et al.\egroup
  }{2020}]{Scala2020SubgoalingTF}
E.~Scala, Patrik Haslum, Sylvie Thi{\'e}baux, and Miquel Ram{\'i}rez.
\newblock Subgoaling techniques for satisficing and optimal numeric planning.
\newblock {\em J. Artif. Intell. Res.}, 68:691--752, 2020.

\bibitem[\protect\citeauthoryear{Seipp and Helmert}{2018}]{cegarjair}
Jendrik Seipp and Malte Helmert.
\newblock Counterexample-guided cartesian abstraction refinement for classical
  planning.
\newblock {\em J. Artif. Int. Res.}, 62(1):535–577, May 2018.

\bibitem[\protect\citeauthoryear{Silver \bgroup \em et al.\egroup
  }{2021}]{TomAndRohanPLOI}
Tom Silver, Rohan Chitnis, Aidan Curtis, Joshua Tenenbaum, Tomas Lozano-Perez,
  and Leslie~Pack Kaelbling.
\newblock Planning with learned object importance in large problem instances
  using graph neural networks.
\newblock In {\em AAAI}, pages 11962--11971, 2021.

\bibitem[\protect\citeauthoryear{Torralba and
  Hoffmann}{2015}]{torralba2015simulation}
Alvaro Torralba and J{\"o}rg Hoffmann.
\newblock Simulation-based admissible dominance pruning.
\newblock In {\em IJCAI}, page 1689–1695, 2015.

\bibitem[\protect\citeauthoryear{Torralba and
  Kissmann}{2015}]{torralbaFocusing}
{\'A}lvaro Torralba and Peter Kissmann.
\newblock Focusing on what really matters: Irrelevance pruning in
  merge-and-shrink.
\newblock In {\em SOCS}, pages 122--130, 2015.

\bibitem[\protect\citeauthoryear{Vallati and
  Chrpa}{2019}]{VallatiPlanningRobustness}
Mauro Vallati and Luk\'{a}\v{s} Chrpa.
\newblock On the robustness of domain-independent planning engines: The impact
  of poorly-engineered knowledge.
\newblock In {\em K-CAP}, page 197–204, 2019.

\bibitem[\protect\citeauthoryear{Vallati \bgroup \em et al.\egroup
  }{2015}]{IPC2014}
Mauro Vallati, Lukas Chrpa, Marek Grze{\'s}, Thomas~Leo McCluskey, Mark
  Roberts, Scott Sanner, et~al.
\newblock The 2014 international planning competition: Progress and trends.
\newblock {\em AI Magazine}, 36(3):90--98, 2015.

\end{thebibliography}

\appendix
\section{Results using Fast Downward's Merge and Shrink Heuristic}

The full configuration we used for the merge and shrink heuristic in table~\ref{table:fd_experiments_ms} is \\


\texttt{astar(merge\_and\_shrink(shrink\_strategy=\\
shrink\_bisimulation(greedy=false),\\
merge\_strategy=merge\_sccs\\(order\_of\_sccs=topological,merge\_selector=\\
score\_based\_filtering\\
(scoring\_functions[goal\_relevance,dfp,\\
total\_order])),label\_reduction=\\
exact(before\_shrinking=true,\\
before\_merging=false),max\_states=50k,\\
threshold\_before\_merge=1))
}




\begin{table*}[t]
\resizebox{\textwidth}{!}{%
\begin{tabular}{lrrrrrrrrrrrrr}
\toprule
\textbf{Problem} & \multicolumn{2}{c}{\textbf{Operators}} & \multicolumn{2}{c}{\textbf{Expansions}} & \multicolumn{2}{c}{\textbf{Evaluations}} & \textbf{Translate} & \textbf{Scoping} & \multicolumn{2}{c}{\textbf{Planning Time (s)}} & \multicolumn{2}{c}{\textbf{Total Time (s)}}\\
  & Unscoped & Scoped & Unscoped & Scoped & Unscoped & Scoped &  &  & Unscoped & Scoped & Unscoped & Scoped\\
\midrule
Driverlog 15 & $2,592$ & $\mathbf{2,112}$ & $527,636$ & $\mathbf{460,244}$ & $8,796,110$ & $\mathbf{7,289,831}$ & $0.5 \pm 0.0$ & $3.6 \pm 0.2$ & $12.4 \pm 0.5$ & $\mathbf{9.4 \pm 0.4}$ & $\mathbf{12.9 \pm 0.5}$ & $13.5 \pm 0.5$\\
Driverlog 16 & $4,890$ & $\mathbf{3,540}$ & $38,681$ & $\mathbf{29,618}$ & $925,264$ & $\mathbf{577,118}$ & $0.7 \pm 0.0$ & $8.4 \pm 0.4$ & $7.6 \pm 0.2$ & $\mathbf{4.5 \pm 0.2}$ & $\mathbf{8.3 \pm 0.3}$ & $13.6 \pm 0.6$\\
Driverlog 17 & $6,170$ & $\mathbf{3,770}$ & $7,768,684$ & $\mathbf{4,607,258}$ & $198,623,900$ & $\mathbf{88,705,990}$ & $0.8 \pm 0.0$ & $9.5 \pm 0.3$ & $154.0 \pm 4.8$ & $\mathbf{50.2 \pm 1.5}$ & $154.9 \pm 4.9$ & $\mathbf{60.6 \pm 1.8}$\\
Logistics 15 & $650$ & $\mathbf{250}$ & $5,951,997$ & $\mathbf{672,736}$ & $140,607,200$ & $\mathbf{11,646,320}$ & $0.3 \pm 0.0$ & $0.7 \pm 0.0$ & $73.7 \pm 2.9$ & $\mathbf{11.1 \pm 0.8}$ & $74.0 \pm 2.9$ & $\mathbf{12.1 \pm 0.8}$\\
Logistics 20 & $650$ & $\mathbf{250}$ & $289,584$ & $\mathbf{86,663}$ & $6,941,424$ & $\mathbf{1,524,997}$ & $0.3 \pm 0.0$ & $0.7 \pm 0.0$ & $\mathbf{7.8 \pm 0.4}$ & $8.6 \pm 0.5$ & $\mathbf{8.1 \pm 0.4}$ & $9.5 \pm 0.6$\\
Logistics 25 & $650$ & $\mathbf{290}$ & $11,437,240$ & $\mathbf{1,944,960}$ & $265,871,100$ & $\mathbf{35,198,000}$ & $0.3 \pm 0.0$ & $0.8 \pm 0.0$ & $148.0 \pm 5.2$ & $\mathbf{21.5 \pm 1.1}$ & $148.3 \pm 5.3$ & $\mathbf{22.5 \pm 1.1}$\\
Satellite 05 & $609$ & $\mathbf{339}$ & $114$ & $114$ & $7,001$ & $\mathbf{3,950}$ & $0.3 \pm 0.1$ & $0.6 \pm 0.1$ & $\mathbf{3.9 \pm 0.2}$ & $4.5 \pm 0.2$ & $\mathbf{4.2 \pm 0.2}$ & $5.5 \pm 0.3$\\
Satellite 06 & $582$ & $\mathbf{362}$ & $21$ & $21$ & $1,084$ & $\mathbf{684}$ & $0.3 \pm 0.1$ & $0.5 \pm 0.1$ & $\mathbf{1.0 \pm 0.1}$ & $1.4 \pm 0.1$ & $\mathbf{1.4 \pm 0.1}$ & $2.2 \pm 0.1$\\
Satellite 07 & $983$ & $\mathbf{587}$ & $> 21,423,400$ & $\mathbf{13,438,440}$ & $> 362,395,000$ & $\mathbf{679,734,000}$ & $0.4 \pm 0.1$ & $0.9 \pm 0.1$ & $> 813.3 \pm 4.2$ & $\mathbf{203.2 \pm 5.6}$ & $> 838.1 \pm 18.3$ & $\mathbf{204.5 \pm 5.7}$\\
Zenotravel 10 & $1,155$ & $\mathbf{1,095}$ & $1,466,718$ & $\mathbf{1,031,280}$ & $37,782,140$ & $\mathbf{25,677,010}$ & $0.3 \pm 0.0$ & $2.4 \pm 0.0$ & $22.1 \pm 0.4$ & $\mathbf{16.6 \pm 0.3}$ & $22.5 \pm 0.4$ & $\mathbf{19.3 \pm 0.4}$\\
Zenotravel 12 & $3,375$ & $\mathbf{3,159}$ & $5,306,442$ & $\mathbf{3,348,866}$ & $231,592,500$ & $\mathbf{140,117,800}$ & $0.6 \pm 0.0$ & $7.3 \pm 0.1$ & $125.4 \pm 4.9$ & $\mathbf{72.3 \pm 3.6}$ & $125.9 \pm 5.0$ & $\mathbf{80.2 \pm 3.7}$\\
Zenotravel 14 & $6,700$ & $\mathbf{6,200}$ & $> 6,462,279$ & $> 5,392,118$ & $> 168,064,648$ & $> 130,752,144$ & $1.0 \pm 0.0$ & $14.3 \pm 0.3$ & $> 644.6 \pm 11.5$ & $> 661.4 \pm 17.0$ & $> 644.6 \pm 11.5$ & $> 676.2 \pm 17.1$\\
\bottomrule
\end{tabular}
}
\caption{Results for our Fast Downward experiments using the Merge and Shrink heuristic. Entries beginning with $>$ indicate that Fast Downward did not find a plan, due to an out-of-memory error. Note that Satellite 07 could only be completed when scoped, and that Logistics 25 was over 6 times as fast when using scoping.}
\label{table:fd_experiments_ms}
\end{table*}

\section{Detailed Experimental Domain Descriptions}
\label{appendix:experiment-details}
Below, we provide detailed descriptions of our different experimental environments.
The PDDL files that were actually used to run these domains in our experiments are included with the provided code submission.
\subsection{Numeric Planning Domains}
\subsubsection{Multi-Switch Continuous Playroom}
In this domain, an agent controls 3 effectors (an eye, a hand, and a marker) to interact with 6 kinds of objects (a light switch, a red button, a green button, a ball, a bell, and a monkey). The  agent exists in a grid where it can take an action to move its effectors in the cardinal directions. To interact with the light switch or buttons, the agent's eye and hand effectors must be at the same grid cell as the relevant object. The light switch can be turned on and off to toggle the playroom's lights, and, when the lights are on, the green button can be pushed to turn on music, while the red button can be pushed to turn off music. Once the music is on, regardless of the state of the lights, the agent can move its eye and hand effectors to the ball and its marker effector to the bell to throw the ball at the bell and frighten the monkey. For this particular goal, if the green buttons are already pressed in the initial state, then all green buttons, as well as all light switches, are rendered task irrelevant.

In our experiments, we created tasks within progressively larger versions of the domain by progressively increasing the number of pressed green buttons and light-switches. The optimal plan for each of these tasks is for the agent to navigate its eye and hand effectors to the ball, and navigate the marker to the bell and then throw the ball at the bell to make the monkey scream.

\subsubsection{Composite IPC Domain}
\label{appendix:composite-ipc}
This domain was constructed by simply combining the numeric \textit{Satellites}, \textit{Driverlog}, \textit{Depots}, and \textit{Zenotravel} domains from the 2002 IPC.
The combination was done straightforwardly: the corresponding sections of the different domain files (i.e. \texttt{types}, \texttt{predicates}, \texttt{functions}, \texttt{operators}) were simply appended to form one combined section in the new composite domain file.
To create the composite problem files, we chose one problem file from each domain and combined the objects and initial states.
However, we did not combine the goals (i.e., we created four separate files that had the same composite objects and initial state, but had the goal of the corresponding original problem file from each of the four respective domains).
Below, we provide a description of the dynamics of the individual domains.
The dynamics of the composite domain are simply the union of those of all the individual domains.

\paragraph{Satellites} In this domain, the agent controls a host of satellites, each equipped with various instruments. The various instruments can be switched on or off independently. They can also be calibrated by pointing them at specific calibration targets. The instruments can also take images of specific phenomena if they are calibrated. Finally, the satellite itself can be angled to point at specific phenomena. Repositioning satellites requires power, and taking readings uses up storage space, both of which are finite. Problems involve procuring images of specific phenomena with specific instruments and pointing specific satellites in particular directions.

\paragraph{Driverlog} In the DriverLog domain, the agent must organize drivers and trucks to transport packages to specific locations. Trucks can be loaded and unloaded with packages, drivers can disembark and board other trucks, and trucks can drive only between locations that are connected. Additionally, driving or walking between locations incurs different amounts of time. Problems involve moving certain drivers, trucks and packages to specific locations with no constraints on time.

\paragraph{Depots} In the Depots domain, the agent must controls various vehicles and cranes that must coordinate together to transport large containers from one location to another. Containers have weights and vehicles have defined load limits that cannot be exceeded. Problems involve ensuring particular containers are left in specific locations.

\paragraph{Zenotravel} In the ZenoTravel domain, the agent must route people and airplanes to specific cities. People can board or disembark from airplanes, and planes can fly between connected cities. When flying, planes can either choose to travel at a normal speed or `zoom', which consumes more fuel. However, zooming can only be done when the number of passengers on the plane is smaller than a prescribed amount. Moreover, planes can be refueled up to a defined capacity. Aircrafts can also be refueled at any location. Most problems require the agent to get various people and planes to  specific cities.

\subsubsection{Minecraft}
In this domain, the agent controls Minecraft's central playable character and can move infinitely in either the x or y directions (though all the interactable objects necessary to complete any of the agent's tasks are located in a fairly small grid in front of the agent as pictured in Figure 1). The agent possesses a diamond axe and three "blocks" of wool in the initial state, and is standing in front of a grid of plants (white flowers, oak saplings and blue flowers). The agent can "pluck" any plant by hitting it repeatedly and then replant it at any location. There is also a set of items, namely four diamonds and seven sticks, beside the grid of plants. The agent can "pick" any of these items by moving to the same location as them, and also place them at any other location. Finally, there are two wooden blocks placed ahead of the grid of plants. Unlike the plants or items, these blocks are solid objects (like the wool blocks the agent already possesses) and will obstruct the agent's path. However, the agent can destroy these blocks with its axe, which them to be automatically picked up by the agent. As with any item, the agent can then place them anywhere, whereupon they will become solid objects again.

The items within the domain can be used to "craft" various other items. If the agent possesses three blue flowers (obtained by plucking), it can invoke an action to craft a blue dye. This dye can be applied to any of the wool blocks to turn them blue. The agent can also craft a diamond axe from three diamonds and two sticks. For every wooden block the agent possesses, it can choose to craft four wooden plank blocks. Finally, the agent can craft a blue bed from three blue-dyed woolen blocks and 3 planks. Note that all these crafted items (except for the diamond axe) are items and can be picked and placed at any location. The bed and wooden planks are solid object blocks that obstruct the agent's movement and must be destroyed with the axe to be picked up and moved.

Within this domain, we defined three different tasks: (1) dye three wool blocks blue, (2) mine wood using a diamond axe and use this to craft wooden planks, and (3) craft a blue bed and place it at a specific location. To complete (1), the agent must pluck the three blue flowers from the center of the grid of plants, craft blue dye, then apply the dye to the wool blocks. To complete (2), the agent must use its axe to break one of the two wooden blocks in the domain, then invoke an action to craft planks. To accomplish (3), the agent must accomplish (1) and (2), then use three dyed wool blocks and three wooden planks to craft a bed. For task (1), the diamond axe sticks and other flowers are causally-unconnected irrelevant. For task (2), the diamond axe is causally-linked irrelevant, the flowers are causally-unconnected irrelevant, and the sticks and diamonds are causally-masked irrelevant. For task (3), the wool blocks and the diamond axe are causally-linked irrelevant, the diamonds and sticks are causally-masked irrelevant, and all plants other than the blue flowers are causally-unconnected irrelevant.

\subsection{IPC Domains}
\label{IPCDomainDescriptions}
Here, we describe the dynamics of each of the planning domains used for Section 5 of the main paper. Note that we did not modify the dynamics of the domain whatsoever for our experiments - we only modified specific problem instances to introduce task irrelevance as described below:

\subsubsection{Logistics}
In this well-known planning domain, the agent is tasked with delivering various packages to specific destination locations. To move the packages, the agent can choose to load them into a plane and fly them between locations or load them into a truck and drive them. Trucks can drive between any locations, but airplanes can only fly between locations with airports. This problem is rather similar to the Depots domain described above in Section \ref{appendix:composite-ipc}.

We introduced irrelevance into problems by modifying the goal so that most packages were already at their goal locations in the initial state.

\subsubsection{DriverLog}
This domain is exactly the same as that described in Section \ref{appendix:composite-ipc}, except that there are no time costs incurred while driving or walking between locations.

We introduced irrelevance into problems by modifying the goal so that most conditions were already satisfied in the initial state. 

\subsubsection{Satellite}
This domain is exactly the same as that described in Section \ref{appendix:composite-ipc}, except that satellites do not have power or data storage limits.

We introduced irrelevance into these problems by modifying the goal so that several of its conditions were already satisfied or almost satisfied (e.g. specific instruments were already calibrated and pointing at goal phenomena) in the initial state. Note that this introduced causally-masked irrelevance by consequence. We also added additional instruments and satellites but not specifying any goal conditions involving these, rendering these irrelevant as well.

\subsubsection{ZenoTravel}
This domain is exactly the same as that described in Section \ref{appendix:composite-ipc}, except there is no limit on the number of passengers that the plane can zoom with, or on the amount of fuel a plane can hold.

We introduced irrelevance into problems by modifying the goal so that many people and planes were already at their goal locations in the initial state.

\subsubsection{Gripper}
In this simple domain, the agent controls a number of grippers. Each gripper can pick up a certain object (but only one at a time), and move between locations while either holding an object or not. Problems typically involve transporting objects between specific rooms.

We did not modify any problems to increase the amount of irrelevance for this domain. Certain problems featured naturally-occurring irrelevance in the form of objects that are not mentioned in the goal whatsoever or objects that are already at their goal.

\end{document}